\documentclass[letterpaper, 10 pt, conference]{ieeeconf}  % Comment this line out if you need a4paper

\IEEEoverridecommandlockouts                              % This command is only needed if 
\usepackage[utf8]{inputenc}
\usepackage{amsmath}
\usepackage{svg}
\usepackage{amssymb} 
\usepackage{graphicx}
\usepackage[colorlinks=true, allcolors=blue]{hyperref}
\usepackage[caption=false]{subfig}
\usepackage{times} % assumes new font selection scheme installed
\usepackage{booktabs}
\usepackage{multirow}
\usepackage{comment}
\let\labelindent\relax
\usepackage{enumitem}
\usepackage{threeparttable}
\usepackage{xurl}
\usepackage{xcolor}

\title{\LARGE \bf
Metrics vs Surveys: An Analysis for Human-Aligned Benchmarking in Social Robot Navigation
}

\author{Stefano Trepella$^{1}$, Mauro Martini$^{1}$, Andrea Ostuni$^{1}$,\\ No\'e P\'erez-Higueras$^{2}$, Fernando Caballero$^{2}$, Luis Merino$^{2}$, and Marcello Chiaberge$^{1}$  % <-this % stops a space
\author{}
\thanks{This work was partially supported by the SWIch action (P.R. F.E.S.R.2021/27 - D.G.R. n.19-6962) within the EMPATHY project, and by PoliTO Interdepartmental Centre for Service Robotics (PIC4SeR). It is also partially supported by the Technology Exchange Programme of the EuROBIN project (GA 101070596) funded by the European Union,
and the project COBUILD (PID2024-161069OB-C31), funded by the Spanish Ministry of Science, Innovation and Universities, the Spanish Research Agency (MICIU/AEI/10.13039/501100011033) and by ERDF "A way of making Europe".} 
\thanks{$^{1}$Department of Electronics and Telecommunications, Politecnico di Torino, 10129, Torino, Italy.
{\tt\footnotesize stefano.trepella@polito.it,  mauro.martini@polito.it, andrea.ostuni@polito.it, marcello.chiaberge@polito.it}}%
\thanks{$^{2}$ School of Engineering, Pablo de Olavide University, Avenida Rectora Rosario Valpuesta, 1. Seville, Spain
        {\tt\footnotesize noeperez@upo.es, fcaballero@upo.es, lmercab@upo.es}}%
}

\begin{document}

\maketitle
\thispagestyle{empty}
\pagestyle{empty}

%%%%%%%%%%%%%%%%%%%%%%%%%%%%%%%%%%%%%%%%%%%%%%%%%%%%%%%%%%%%%%%%%%%%%%%%%%%%%%%%
\begin{abstract}

Social, also called human-aware, navigation is a key challenge for integrating mobile robots into human environments. The evaluation of such systems is complex, as factors such as comfort, safety, and legibility must be considered. Human-centered assessments, typically conducted through surveys, provide reliable insights but are costly, resource-intensive, and difficult to reproduce or compare across systems. Alternatively, numerical social navigation metrics are easy to compute and facilitate comparisons, yet the community lacks consensus on a standard set of metrics.

This work explores the relationship between numerical metrics and human-centered evaluations to identify potential correlations. If specific quantitative measures align with human perceptions, they could serve as preliminary benchmarking tools, providing a human-aligned assessment when large-scale surveys are not feasible. Our results indicate that while current metrics capture some aspects of robot navigation behavior, important subjective factors remain insufficiently represented, necessitating new metrics.

\end{abstract}

%%%%%%%%%%%%%%%%%%%%%%%%%%%%%%%%%%%%%%%%%%%%%%%%%%%%%%%%%%%%
% INTRODUCTION
%%%%%%%%%%%%%%%%%%%%%%%%%%%%%%%%%%%%%%%%%%%%%%%%%%%%%%%%%%%%
\section{Introduction}
\label{sec:intro}
Human-aware robot navigation is a key research area for integrating mobile robots into human environments \cite{moller2021survey, eirale2025human}. Beyond the classical challenges of path planning and obstacle avoidance, human-aware navigation must address qualitative aspects of social interaction, such as comfort, predictability, and personal space, which are difficult to capture with mathematical models \cite{mavrogiannis2023core, Francis2025}.

Precisely, these qualitative aspects make the evaluation of human-aware navigation complex. In addition to traditional performance metrics (e.g. time, distance), several quantitative metrics for social navigation have been proposed \cite{gao2022evaluation, wang2022metrics}. These metrics are easy to compute in controlled experiments and enable efficient comparisons between approaches. However, adopting a wide set of contrastive measures is often misleading in clearly understanding the quality of an experiment. Furthermore, there is no agreement on a set of metrics that reliably capture social aspects such as comfort or legibility. Such aspects are typically assessed through participant surveys, which, while informative, are costly, time-consuming, and challenging to scale.

\begin{figure}
    \centering
    \includegraphics[width=\linewidth]{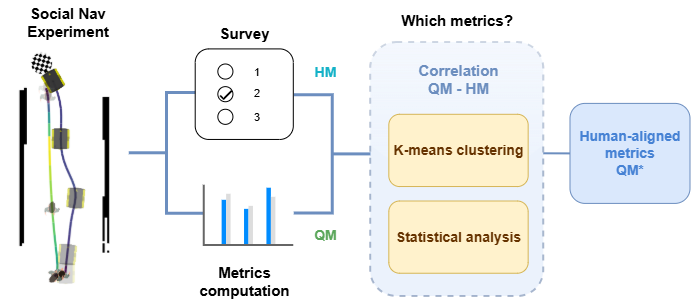}
    \caption{The proposed framework to infer human metrics (HM) for social navigation experiment using an optimal set of representative quantitative metrics QM*. From the set of experiments, we obtain quantitative metrics from the trajectories and human-provided survey metrics. Then, we use a statistical analysis to determine the correlation between QM and HM, and the optimal set QM*.}    \label{fig:first_page}
\end{figure}

This work investigates
whether it is possible to infer qualitative aspects from quantitative metrics measured in experiments. We approach the problem by analyzing correlations between quantitative evaluations and qualitative human judgments from surveys in experiments with real robots and human participants. Our goal is to identify quantitative metrics that align with human perceptions, enabling standardized quantitative evaluations that can supplement human-centered validation and provide consistent preliminary benchmarks. A schematic of the proposed framework is illustrated in Fig. \ref{fig:first_page}. The correlation analysis combines K-means clustering with statistical methods to identify sets of quantitative metrics strongly related to human evaluations. 

The contributions of this work can therefore be summarized as follows:

\begin{enumerate}
    \item A novel framework to perform a joint analysis of surveys and quantitative metrics, offering insights on the correlations between them through clustering and statistical tests.
    \item The identification of a subset of quantitative metrics that results in the most relevant human-like evaluation trends for social navigation experiments.
    \item A dataset of robot and people trajectories in eight different social scenarios collected through real-world experiments with an accurate ground truth and survey evaluation, that can be used for development and benchmarking of future research.
    % \item A clustering approach to identify relevant sets of quantitative metrics matching with human-centered evaluation of social navigation.
    % \item A statistical analysis of correlations between individual quantitative metrics and human evaluations.
\end{enumerate}

The correlation analysis framework, the dataset, and the survey questionnaire are available in the repository\footnote{\url{https://github.com/PIC4SeR/Social-Nav-Metrics-Matching}}.

%%%%%%%%%%%%%%%%%%%%%%%%%%%%%%%%%%%%%%%%%%%%%%%%%%%%%%%%%%%%
% RELATED WORK
%%%%%%%%%%%%%%%%%%%%%%%%%%%%%%%%%%%%%%%%%%%%%%%%%%%%%%%%%%%%
\section{Related Work}
\label{sec:related_works}

In the last decade, several studies have made significant contributions to the definition and evaluation of social navigation in robotics. The studies of Francis et al. \cite{Francis2025}, or Mavrogiannis et at. \cite{mavrogiannis2023core} make a thorough analysis of the challenges in evaluating social navigation algorithms. Another example is the work of Gao and Huang \cite{gao2022evaluation}, which provides an in-depth review of methods that quantify a robot’s adherence to social norms. Their study examines metrics such as maintaining appropriate interpersonal distances, minimizing discomfort among human co-habitants, and ensuring that paths are socially acceptable.  
% usual set of quantitative metrics
Classical quantitative benchmarks are usually derived from the analysis of the Proxemics of human-human interaction distances \cite{proxemics_Hall}, as in the review in \cite{samarakoon2022review}. Furthermore, other theories of social navigation behavior, based on the famous Social Force Model (SFM) \cite{helbing1995social}, allow quantification of the social impact of robot motion using Social Forces and Social Work, taking into account agents' velocities and directions \cite{Martini2024-SFWP}.

In recent years, a new wave of robotics simulators emerged with improved human motion and benchmarks for human-aware navigation: HuNavSim \cite{PerezRal2023}, SEAN2.0 \cite{sean2_Tsoi_RAL22}, SocNavBench \cite{biswas2022socnavbench}, Arena 4.0 \cite{Arena4_2024}. These simulators exploit trajectory ground truth and repeatability facilities to compute quantitative metrics for both classical and social navigation.  

% New trends in evaluating social navigation experiments
The most recent research promotes adopting new, comprehensive metrics that integrate efficiency, safety, and human perception into a single score. Among them \cite{wang2022metrics}, \cite{karwowski23metrics}, and \cite{singamaneni2023towards} propose a suite of novel metrics that aim to capture human-centric perceptions. The latter evaluates robot behavior from the perspective of human comfort and social acceptability, combining objective measurements with qualitative data based on surveys. This dual perspective allows researchers to better analyze whether conventional metrics completely capture critical aspects of human-robot interaction.
In addition to defining new metrics, a recent study also highlighted the importance of assigning the correct level of complexity to social scenarios \cite{stratton2024characterizing}.

In general, the most recent studies highlight a remaining challenge: aligning quantitative benchmarks with more subjective human evaluations of robot navigation in social contexts. This work analyzes the correlation between currently used quantitative metrics and survey-based human assessment, posing the question whether these hidden patterns can lead to a new evaluation protocol that integrates human feedback with algorithmic metrics.

%%%%%%%%%%%%%%%%%%%%%%%%%%%%%%%%%%%%%%%%%%%%%%%%%%%%%%%%%%%%
% DATA COLLECTION
%%%%%%%%%%%%%%%%%%%%%%%%%%%%%%%%%%%%%%%%%%%%%%%%%%%%%%%%%%%%
\section{Data collection}
\label{sec:data_collection}
In this section, we describe the steps conducted to collect the social navigation dataset and frame the analysis in a precise evaluation protocol. First, we detail the settings of the experimental activity carried out to record trajectories of robot and people moving in the same environment. Then, the eight social scenarios are defined depicting the human motion, the robot motion, and the maps. Quantitative and human-centric metrics considered for the evaluation are therefore defined, and the statistics and format of the survey activity are finally reported.

\subsection{Experimental setup}
\label{subsec:exp_settings}

\begin{figure}[t]
    \centering
    \includegraphics[width=0.95\linewidth]{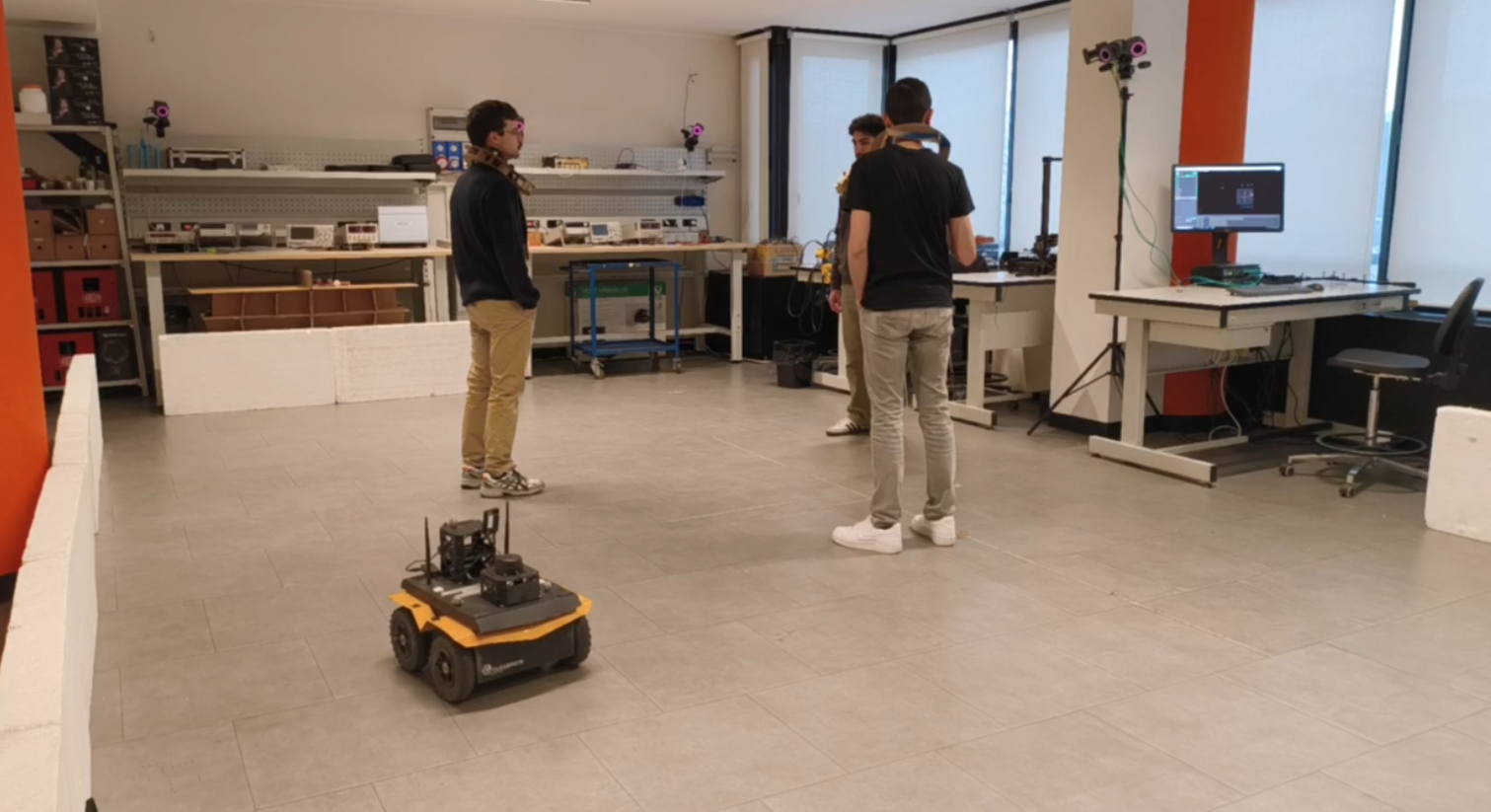}
    \caption{The laboratory where experiments were performed using a Jackal robot and a VICON tracking system to collect trajectories.}
    \label{fig:jackal_in_lab}
\end{figure}

Experimental data have been collected using a real robot in laboratory spaces equipped with a VICON tracking system to record the trajectories of both the robot and humans. The robot adopted is a Jackal from ClearPath Robotics, a typical skid-steer drive robotic platform. An example picture taken in the lab during the experiments is shown in Fig. \ref{fig:jackal_in_lab}. The effective area used in the laboratory is $4.0  \times 5.0$ m, with the first two maps being $2.0 \times 5.0$ m, the third being $4.0 \times 3.0$ m, and the fourth being $3.0 \times 4.0$ m.
In the experiments, the robot autonomously navigates using different local planning algorithms, resulting in behaviors with varying levels of social compliance. The controllers have been tuned for each scenario to achieve smoother paths or a less conservative policy, resulting in greater agility and responsiveness.
Moreover, in some experiments, the controllers have been coupled with an additional social cost term to exhibit socially compliant trajectories. For instance, a Gaussian costmap around people and the Social Work term defined in Section \ref{subsec:metrics} were used as cost terms in the control function to guide the computation of the robot's velocity commands.
The velocity limits of the robot have been set to $v \in [0.0, 0.6]$ m/s and $\omega\in [-1.5, 1.5]$ rad/s for all experiments and the different controllers used.

\begin{figure*}[!t]
    \centering
    \subfloat[Map 1]{
       \includegraphics[width=0.23\linewidth]{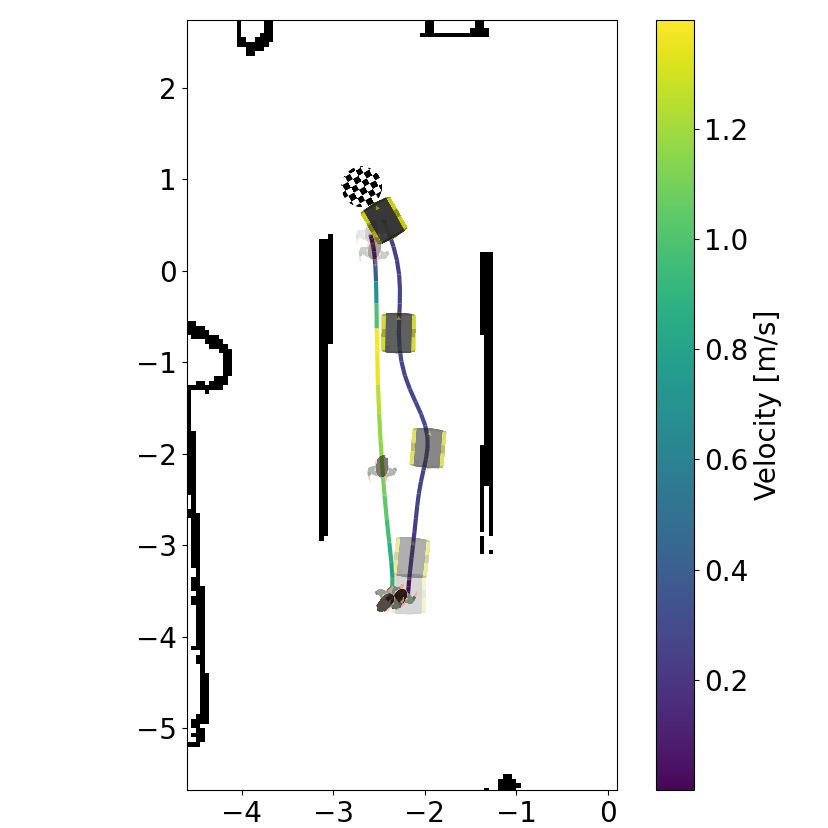}
       \label{fig:passing}
    }
    %\hfill
    \subfloat[Map 2]{
       \includegraphics[width=0.23\linewidth]{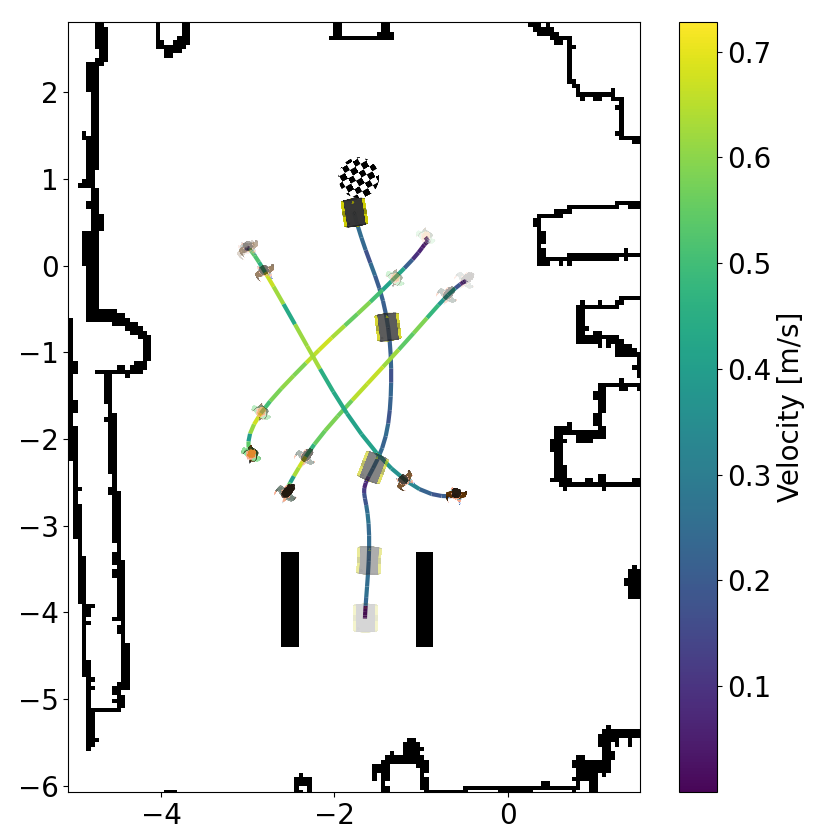}
       \label{fig:crossing_3}
    }
    %\hfill
    \subfloat[Map 3]{
       \includegraphics[width=0.23\linewidth]{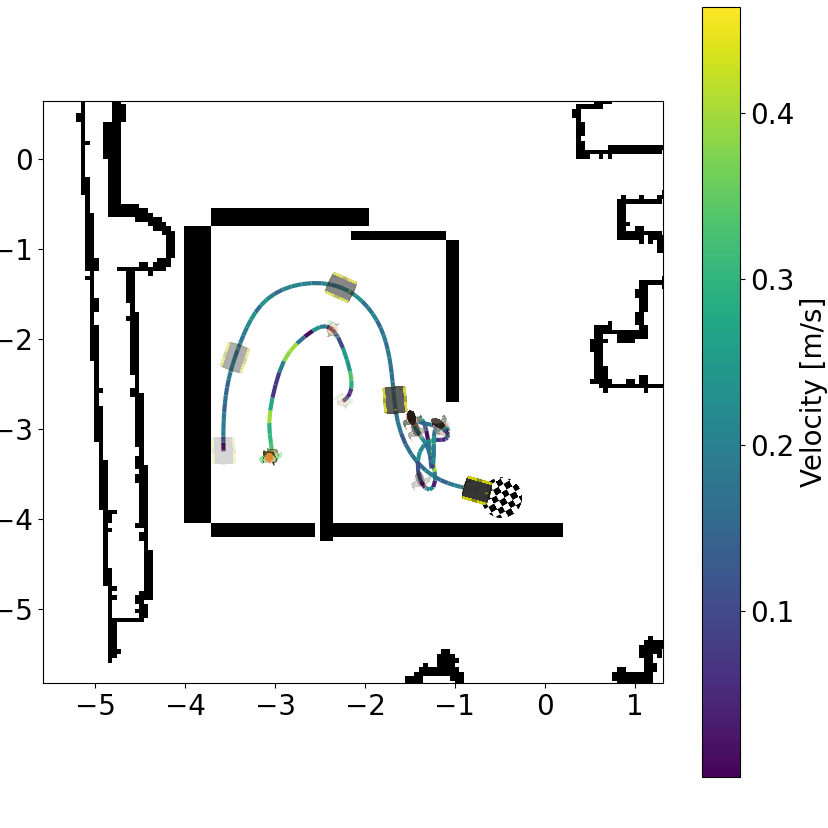}
       \label{fig:narrow_turn}
    }
    \subfloat[Map 4]{
       \includegraphics[width=0.23\linewidth]{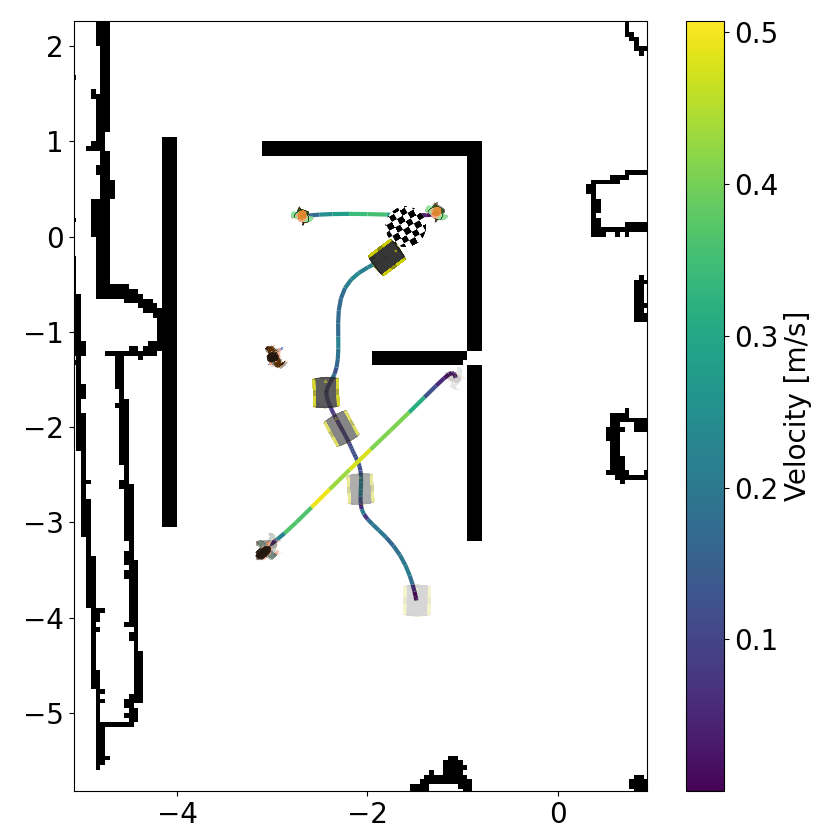}
       \label{fig:mixed}
    }
    \caption{Maps used for the experiments and trajectories of robot and humans in four distinct scenarios: Passing, Crossing 3, Narrow turn and Mixed. The velocity scale is indicated by the colorbar on the side, while temporal evolution is expressed through transparency. Other scenarios differ in human motion, as explained in Section \ref{subsec:social_scenarios}.}
    \label{fig:scenario_maps}
\end{figure*}

\subsection{Social Scenarios}
\label{subsec:social_scenarios}
Eight different social scenarios have been designed to collect trajectories with robots and humans. The scenarios of \textit{Passing} and \textit{Overtaking} were carried out in Map 1 shown in Fig. \ref{fig:passing}. These situations represent two standard interactions in social navigation. Passing involves a robot traversing a corridor while a person walks in the opposite direction, instead overtaking sees the robot trying to surpass a person walking in the same direction. 
Scenarios \textit{Crossing 1, 2, 3} and \textit{Curious person} were conducted in Map 2 of Fig. \ref{fig:crossing_3}. The crossing scenarios involve the robot traveling forward towards its target, while a varying number of people cross its trajectory to reach their destination. Three variations of this case are experimented with: \textit{Crossing 1} has two traversing humans starting from the same side, in \textit{Crossing 2} two people start from opposite sides, and in \textit{Crossing 3} there are three people walking in a cross pattern, two of them being side-by-side. 

Maps 3 and 4 were included to test social navigation in more advanced and hybrid scenarios, respectively, as shown in Figs. \ref{fig:narrow_turn} and \ref{fig:mixed}. In Scenario \textit{Narrow turn}, the robot must perform a U-shape blind turn with a person appearing from behind the corner and then reach the goal while avoiding a person walking in the same narrow passage. Scenario \textit{Mixed} is a complex scenario obtained by combining different situations. The robot traverses a room with three people, one of them standing still, and the last one walking close to the robot's target.
 Lastly, the scenario \textit{Curious person} involves a curious person who actively follows the robot and obstructs its path. 

In the first seven scenarios, the humans walk at a moderate speed, between 0.4 and 1 m/s, treating the robot as another person and modifying their trajectories only to avoid potential collisions. In contrast, in Scenario \textit{Curious person}, the person chases and disrupts the robot's motion, acting as a curious agent.
Three runs of each scenario were performed with different settings of the robot controllers,leading to different social-navigation behaviors.
%The collected trajectories  of the robot and the agents were turned into GIFs to have a top-down view of the experiments. These \textcolor{blue}{2D animations} are shown in the survey for evaluation.

\subsection{Quantitative metrics}
\label{subsec:metrics}
%The metrics used in this study can be categorized into two distinct types: 

%\begin{description}[leftmargin=0pt, font=\normalfont\textbf]

%\item[Quantitative metrics (QM):] 
We employ a set of eleven numerical metrics used in the literature and compiled in standard social navigation benchmarks, such as HuNavSim \cite{PerezRal2023}. This QM set is composed of four metrics focused on classic navigation aspects, and seven metrics for social behavior:

\begin{itemize}
    \item \textit{Time to goal} ($TTG$) [s]
    \item \textit{Path length} ($PL$) [m]
    \item \textit{Cumulative heading changes} ($CHC$) [rad]
    \item \textit{Average robot linear velocity} ($ARV$) [m/s]
    \item \textit{Social work} ($SW$)
    \item \textit{Social work per second} ($SW_{s}$)
    \item \textit{Average minimum distance to closest person} ($AMD$) [m]
    \item \textit{Proxemics occupancy}: intimate space ($PR_I$), personal space ($PR_{PE}$), social space ($PR_S$), public space ($PR_{PU}$).
    
\end{itemize}

$CHC$ represents the total angular distance spanned by the robot during the path. The quantity of social work is derived by the Social Force Model (SFM) of interaction between a crowd of agents proposed in \cite{helbing1995social}. 
The social work $SW$ is computed at each time step for the robot according to the following expression, considering the velocity module and direction of each agent around the robot: $SW = W_r + \sum_i W_{p,i}$, where $W_r$ is the sum of the modulus of the robot social force ($F_P$) and the robot obstacle force ($F_O$) along the trajectory according to the SFM, while $W_p$ is the sum of the modulus of the social forces generated by the robot for each pedestrian $i$ along the trajectory in a maximum range of 5 meters. Social work is included both as an average per second and as an overall sum throughout the duration of the experiment.

The proxemics metrics are four complementary percentages that represent the time spent by the robot in a given zone: closest to the person (intimate), close to the person (personal), a reasonable distance from the person but still in a range of possible interaction (social), completely away from the agent (public).

\subsection{Human metrics from survey Questionnaire}
\label{subsec:survey}

We have employed a set of four human-centric qualitative metrics (HM) to rank the experiments across relevant aspects, as proposed by Hirose et al. \cite{hirose2024selfi}. They are obtained through questionnaires sent to participants using the 5-point Likert scale. These metrics are:

\begin{itemize}
    \item \textit{Unobtrusiveness} ($UO$): the robot's ability to move in the environment without being noticed by humans. 
    \item \textit{Friendliness} ($FL$): the level of social disturbance of the robot in its trajectory.
    \item \textit{Smoothness} ($SM$): the robot's smoothness in its trajectory. 
    \item \textit{Avoidance foresight} ($AF$): the robot's ability to adjust its motion in advance to avoid possible future collisions with moving people.
\end{itemize}

The human metrics (HM) have been collected through a survey based on real robot experiments conducted in the lab. For each required assessment, the social navigation scenario is described, explaining the robot task, the situation from the pedestrians perspective, and the metrics significance. The collected trajectories of the robot and the agents were converted into GIFs to provide a top-down view of the experiments. These 2D animations are shown in the survey for evaluation, as can be seen in the survey questionnaire form \footnote{\url{https://forms.gle/qyKHhzeYnaH8TPvQ7}}. A total of 24 experiments have been evaluated with the survey. Fig. \ref{fig:scenario_maps} shows the four maps used for the experiments and trajectories examples obtained in four different runs of Passing, Crossing 3, Narrow turn, and Mixed scenarios. For the special case of Scenario \textit{Curious person}, the unobtrusiveness score has not been collected in the survey, since the intrusive behavior of the human makes it lose relevance.
It has been shown in prior work \cite{hirose2024selfi} that by using 2D animations, we focus the participant's attention on physical motion signatures such as velocity and relative distances that results to be primary indicators of social intent.
The survey included 70 participants from Italy, Spain, and France, aged 20 to 60 years, with a mean age of 31 years and about 40\% female participants. The background knowledge in robotics and autonomous navigation has been rated low for almost 50\% of people (up to 2 out of 5), and high for about 30\% of people (4 or 5 out of 5).
The participants independently rated the experiments on a 5-point Likert scale to avoid relative-ranking bias across scenarios.

%%%%%%%%%%%%%%%%%%%%%%%%%%%%%%%%%%%%%%%%%%%%%%%%%%%%%%%%%%%%
% ANALYSIS
%%%%%%%%%%%%%%%%%%%%%%%%%%%%%%%%%%%%%%%%%%%%%%%%%%%%%%%%%%%%
\section{Analysis}
\label{sec:analysis}
In this section, we describe the methods adopted to process the data extracted from the social navigation experiments and the surveys. We aim to investigate correlations among metrics and propose a new practical evaluation protocol based solely on quantitative aspects. 

The framework is summarized in Fig. \ref{fig:first_page}. %A total of 24 social navigation experiments have been conducted in real world and evaluated through a survey. 
Metrics correlation with humans scores is investigated to identify the most representative quantitative metrics through a clustering approach, together with a statistical analysis for single-metric correlation. 
\subsection{Metrics Pre-Processing}
Given the different measurement units and numerical ranges of the adopted QM metrics, the metrics were rescaled between 0 and 1. In this way, it is possible to perform a direct numerical comparison of the metrics across experiments with diverse geometric layouts, numbers of humans, and motion control algorithms. For each scenario, the run with the best value of a particular QM is assigned a value of 1, while the other two are assigned a lower linearly scaled value, depending on their share of that particular metric. The definition of best value varies depending on the specific metric used to evaluate navigation efficiency or safety. For example, Social Work is better when minimized since higher values represent a greater social impact or disturbance to the surrounding pedestrians' intended paths, according to the Social Force Model theory
\cite{helbing1995social}. Similarly, based on Hall's proxemic principles \cite{proxemics_Hall}, minimizing occupancy in intimate and personal spaces is prioritized to ensure human comfort and to respect social boundaries. Conversely, social space occupancy is considered better when higher, as it reflects the robot maintaining a socially appropriate but safe distance. Similar considerations were applied to all other QM metrics.
The HM metrics have also been scaled in $[0, 1]$ by dividing by 5, making these data ready for aggregation, comparison, and clustering.

\subsection{Clustering}\label{subsec:cluster}
The QM and HM metrics data represent two distinct feature sets of the same 24-experiment sample. In this section, we investigate the overall similarity between the two metric sets by grouping the experiments using the QM and HM feature spaces separately. While human labels were available, we specifically chose an unsupervised K-means clustering approach \cite{likas2003global} to determine if the inherent, high-dimensional structure of the quantitative data naturally aligns with human categorical perceptions without the bias of supervised training. This enables a robust assessment of whether numerical metrics can autonomously capture the underlying patterns of social interaction that humans observe.
The clusters obtained are first analyzed for their internal composition to determine the optimal number of groups, K. Then, clustering accuracy is used to assess whether the QM and HM feature spaces yield the same experimental groups.
The internal composition analysis is performed employing the silhouette score, a metric that quantifies the coherence of clusters. %The coefficient, which ranges from -1 to 1, shows that values approaching 1 indicate that clusters are clearly dissimilar from each other, while negative scores imply the potential for incorrect cluster assignment for certain samples.
The K values tried to group the 24 experimental samples are in the range $[2, 5]$.

The matching alignment of the clustering is evaluated using the Adjusted Rand Index (ARI) \cite{steinley2004properties}, which is frequently used in data clustering analysis. ARI is a coefficient that quantifies the similarity between two divisions of a dataset by comparing how data points are assigned to clusters. For our analysis, the labels assigned to the experiments by clustering using the HM survey metrics were treated as the ground truth. 
The ARI similarity is computed for the entire set of quantitative metrics and for every possible combination of QMs, except the empty set. This process yielded a total of $2.047$ combinations.
The objective of this analysis is to identify the most prevalent metrics in the combinations that produce the highest ARI. For this purpose, a cumulative ARI score has been computed for each metric, summing the ARI values obtained over the entire QM combination. 

The final output, i.e., the cumulative ARI for each metric, represents the relevance of these metrics in achieving the same groupings obtained with the HM features space using K-means clustering.

\subsection{Statistical Methods for Evaluating Metrics}

%The correlation between QM and HM sets of metrics is estimated by adopting two non-parametric statistical methods: Spearman's Rho and Kendall's Tau. Both approaches are ideal for data that do not necessarily meet the assumptions of normality, making them well-suited for ordinal or rank-based evaluations standard in social navigation studies. Combining these approaches not only provides complementary views of the correlation structure but also enables a comprehensive evaluation of how well the quantitative metrics capture trends observed in human assessments.

The correlation between QM and HM metrics is evaluated using three non-parametric statistical methods: Spearman’s Rho, Kendall’s Tau, and the Kruskal–Wallis H-test. These methods do not assume normal data distributions and are therefore well suited for ordinal or rank-based evaluations common in social navigation studies.

\begin{description}[leftmargin=0pt, font=\normalfont\textbf]

\item[Spearman:] Spearman’s Rho evaluates monotonic relationships by converting metric values into ranks and measuring the correlation between paired rank sets.

\item[Kendall:] Kendall’s Tau measures association by comparing concordant and discordant pairs of ranked observations. The coefficient reflects the proportion of agreement between the orderings of two variables, making it particularly robust for smaller datasets.

\item[Kruskal-Wallis:] The Kruskal–Wallis H-test evaluates whether multiple independent groups originate from the same distribution. It compares the rank distributions across groups and determines whether at least one group differs significantly.

\end{description}

%\subsection{Overall metrics matching}
%As a final analysis, to gain practical insight from the correlation study, an aggregate score is calculated for HM and QM metrics across each social scenario. The overall scores are computed averaging equally the metrics in the HM set, in the initial complete QM set, and in the correlation-driven optimal QM set. Hence, in Section \ref{sec:results}, these aggregated values are directly compared to demonstrate how the correlation analysis conducted can lead to the identification of a minimal set of relevant QM metrics that can be used to evaluate a social navigation experiment preserving the human-level assessment. The goal of this comparison is to demonstrate that the reduced QM set can be used to represent the overall evaluation trends and general ranking obtained through the survey. Thus, the complete proposed study can serve as a benchmarking protocol to evaluate new experiments using only the QM, \textcolor{blue}{offering a more efficient preliminary assessment tool while acknowledging that large-scale human surveys remain the gold standard for final system validation.}

%\subsection{\textcolor{blue}{Statistical methods}}
%ToDo: we should comment here the new methods we must add to validate the trends in Figure \ref{fig:histogram_aggregate_scores}.

%%%%%%%%%%%%%%%%%%%%%%%%%%%%%%%%%%%%%%%%%%%%%%%%%%%%%%%%%%%%
% RESULTS
%%%%%%%%%%%%%%%%%%%%%%%%%%%%%%%%%%%%%%%%%%%%%%%%%%%%%%%%%%%%
\section{Results}
\label{sec:results}
In this section, results are presented and discussed. First, the correlation between metrics and human-centric evaluations is demonstrated through clustering experiments across both the QM and HM feature spaces. Certain subsets of metrics that maximize the ARI between clusters are therefore identified. Then, individual metrics correlation with the HM set is demonstrated through the statistical analysis.
Finally, an overall score comparison demonstrates that quantitative metrics identified through correlation show evaluation trends that align with human assessment.
%An overall comparison is finally performed with aggregated scores, showing the enhanced similarity between correlation-based optimal QM and the reference HM.

\subsection{Statistical Validation of Human Ground Truth (HM)}

\begin{table}[t]
\centering
\caption{Analysis of HM set }
\label{table:hm_analysis}
% \begin{tabular}{lcc}
\begin{tabular}{lcccc}

\toprule
% \textbf{Social Scenario} & \textbf{Friedman} & \textbf{Wilcoxon Sign. rank} \\ \hline
% Passing                  & $<$ 0.001 (***)      & $<$ 0.001 (***)   \\
% Overtaking               & 0.0016 (**)          & 0.0011 (**) \\
% Crossing 1               & $<$ 0.001 (***)      & 0.0480 (*) \\
% Crossing 2               & $<$ 0.001 (***)      & $<$ 0.001 (***)  \\
% Narrow Turn              & $<$ 0.001 (***)      & 0.0213 (*)  \\
% Mixed                    & $<$ 0.001 (***)      & $<$ 0.001 (***) \\
% Crossing 3               & $<$ 0.001 (***)      & 0.9084 (ns)  \\
% Curious                  & 0.0030 (**)          & 0.2947 (ns) \\ \hline

\textbf{Scenario} & \textbf{Friedman} & \textbf{Wilcoxon signed-rank test}\\ \midrule
Passing    & $<0.001$ (***) & $<0.001$ (***)\\
Overtaking & $<0.001$ (***) & $<0.001$ (***)\\
Crossing 1 & $<0.001$ (***) & $<0.001$ (***)\\
Crossing 2 & $<0.001$ (***) & $<0.001$ (***)\\
Narrow Turn & $<0.001$ (***) & 0.0040 (**)\\
Mixed & $<0.001$ (***) & $<0.001$ (***)\\
Crossing 3 & $<0.001$ (***) & $<0.001$ (***)\\
Curious & 0.0025 (**)   & $<0.001$ (***)\\
\bottomrule
\end{tabular}
\begin{tablenotes}
      \footnotesize
      \item Significance levels: *** $p < 0.001$, ** $p < 0.01$, * $p < 0.05$, ns = not significant ($p > 0.05$).
\end{tablenotes}
\end{table}

First, we analyze the HM collected. The aim is to mathematically prove that they can clearly identify the different social behaviors across the different scenarios, as expected. We demonstrate that the human perception data collected in the survey is not merely a set of subjective biases, but a consistent and statistically significant measure capable of distinguishing among varying robot behaviors. By validating these responses, we establish the human survey data as the definitive benchmark against which automated quantitative metrics are calibrated.
The Friedman test \cite{Friedman1937} was employed as an omnibus non-parametric test to assess whether evaluation scores differed significantly across the three experimental runs within each scenario. This analysis determines whether different controller configurations produced perceptible differences in navigation quality.
Following significant results, post-hoc pairwise comparisons were conducted using the Wilcoxon signed-rank test \cite{Wilcoxon1945}. In particular, comparisons were performed between Run~3 (expected to exhibit the most socially compliant behavior) and Run~1 (expected to be the least socially compliant).
This post-hoc analysis allows identifying which specific behaviors were perceived as significantly different and supports the ranking of navigation strategies in terms of perceived social quality.
Table \ref{table:hm_analysis} summarizes the Friedman test results for the aggregated human responses, along with the Wilcoxon results. According to Friedman, the HM survey data maintains a statistically significant score gradient across all eight scenarios ($p<0.01$), establishing a robust ground truth. Interestingly, in the Narrow Turn  scenario, the human participants were statistically slightly less consistent. This suggests that humans find advanced and mixed interactions more complex to evaluate, often expressing a more severe assessment for any trajectory that is only partially social.

\begin{figure}[t]
    \centering
    \includegraphics[width=\linewidth]{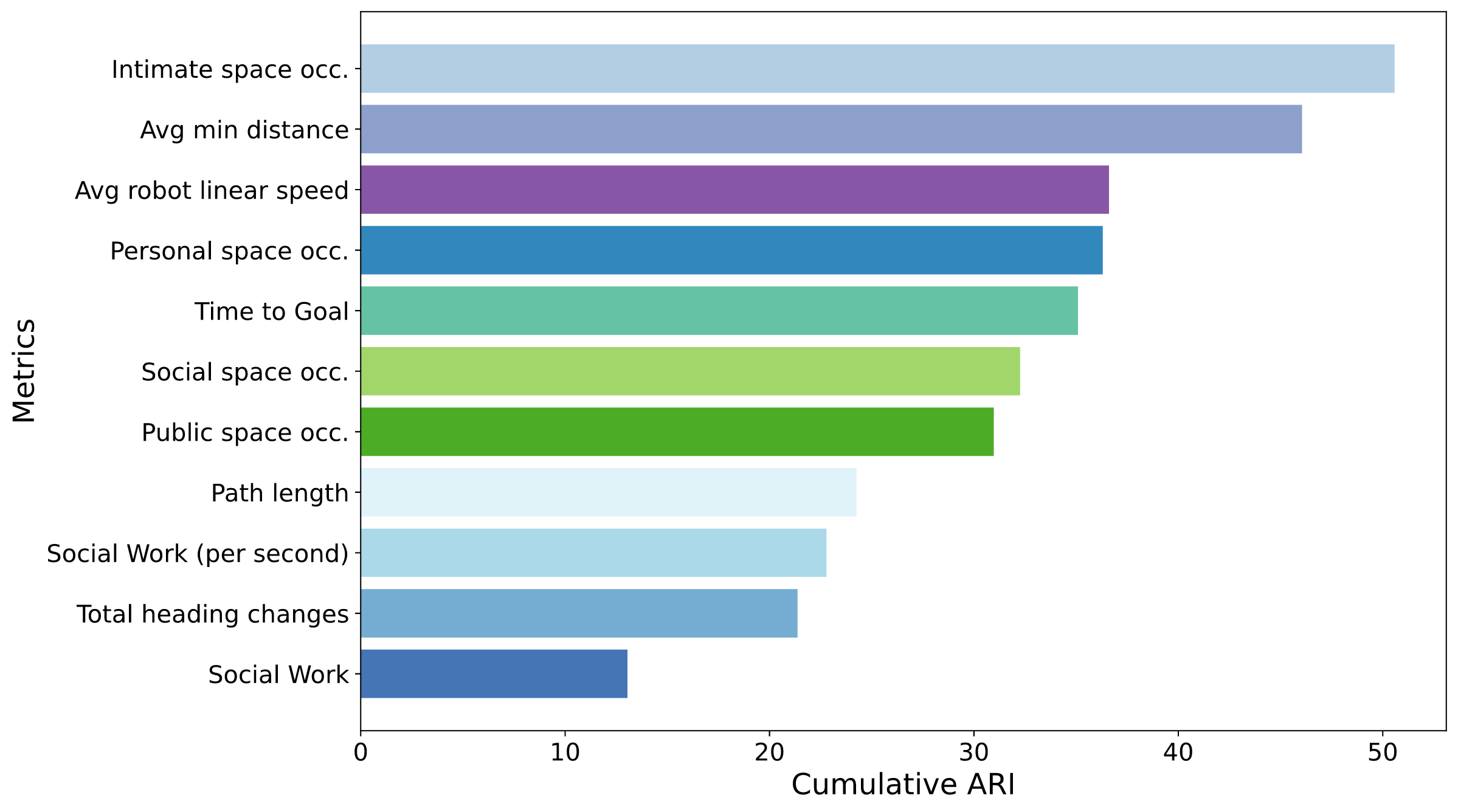}
    \caption{Histogram of the cumulative ARI. The QM metrics with the highest cumulative ARI are more relevant for clustering experiments, as shown using the HM feature set.}
    \label{fig:histogram_cumulative_ARI}
\end{figure}

\subsection{Clustering comparison}
The K-means clustering algorithm is used to study the spatial distribution of experimental samples using QM and HM data as feature spaces. First, we investigate the separability of samples in different numbers of clusters for the HM samples. This internal cluster composition analysis yielded a silhouette score of 0.548 for $K=2$, while $K=3$ yielded 0.435, indicating that division into two clusters is more easily distinguishable.
Then, we aimed to investigate the matching accuracy between clusters of experiments obtained using two different feature spaces, represented by the HM and QM metrics. This study has been conducted for both $K=2,3$, computing a cumulative ARI score as explained in Section \ref{subsec:cluster}.
The results of the clustering accuracy analysis are summarized by the histogram presented in Fig. \ref{fig:histogram_cumulative_ARI}. Regardless of the number of clusters chosen in K-means, the five metrics with the highest cumulative ARI are: intimate space occupancy, average minimum distance to the closest person, personal space occupancy, average robot linear velocity, and time to goal. This distribution ensures a good balance between the social and trajectory's efficiency metrics. Therefore, both of these aspects contribute to a complete human-like evaluation.

In Table \ref{tab:optimal_ARI_combos}, the optimal combinations of QM metrics that maximize the ARI are reported, considering $K=2$. These optimal QM sets significantly increase the ARI compared to using the entire QM set as the feature space. The initial ARI of $-0.04$ across all 11 quantitative metrics was inadequate to establish a substantial correlation between the two clusters, as it was lower than 0.3. 
All optimal QM combinations comprise at least three of the most prevalent metrics identified with the cumulative ARI computation. In particular, the average robot velocity and one of the proxemics area occupancy are always present, highlighting the importance of this information for a quantitative evaluation protocol.

\begin{table}[t]
    \centering
    \caption{The best four QM sets that maximize the ARI scores with K-means clustering using $K=2$.}
    \label{tab:optimal_ARI_combos}
    \begin{tabular}{ccccc}
        \toprule
        \textbf{Initial ARI} & \textbf{ ARI} & \textbf{QM set} \\
        \midrule
         -0.04 & 0.54 & $ARV$, $SW_s$, $PR_I$, $PR_{PE}$, $AMD$ \\
         -0.04 & 0.42 & $TTG$, $ARV$, $PL$, $SW_s$,$PR_I$, $PR_{PE}$ \\
        -0.04 & 0.42  & $ARV$, $PR_S$, $AMD$ \\
         -0.04 & 0.31 & $TTG$, $ARV$, $SW_s$, $PR_{PE}$, $AMD$ \\
        \bottomrule
    \end{tabular}
\end{table}

\subsection{Statistical Analysis of QM and HM Metrics}
\begin{figure}[t]
    \centering
    \includegraphics[width=\linewidth]{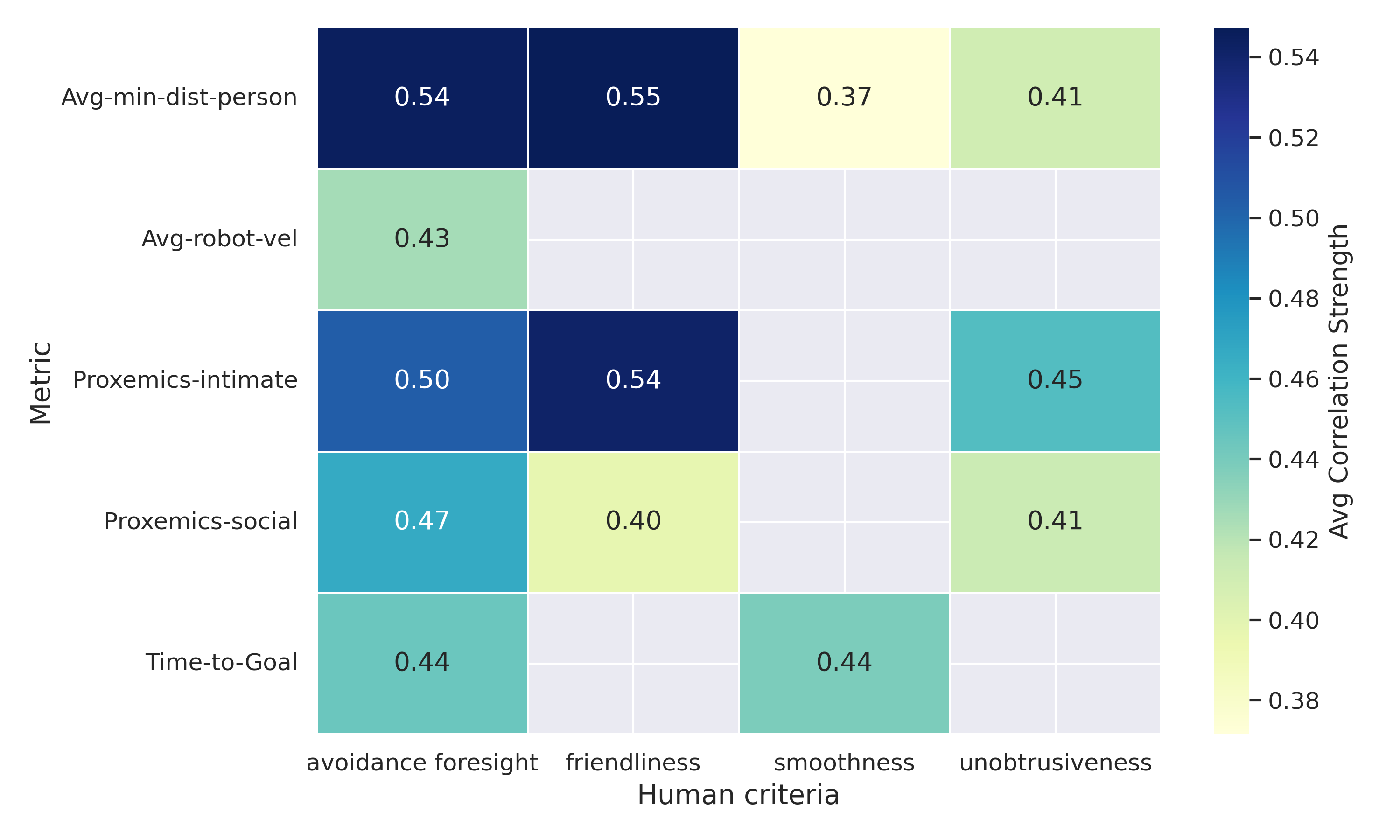}
    \caption{Heatmap of consistent correlations between single QM and HM metrics, according to both Kendall and Spearman analysis.}
    \label{fig:corr_heatmap}
\end{figure}

A statistical analysis has been conducted to investigate the correlation between the QM and HM sets using a single metric, with the aim of identifying the most representative quantitative metric for each human factor considered. First, we use two non-parametric statistical methods: Spearman’s Rho and Kendall’s Tau. Both approaches are ideal for data that do not necessarily meet the assumptions of normality, making them well-suited for the ordinal 5-point Likert-scale evaluations used in our survey. We considered correlations that obtain a Spearman strength $|\rho > 0.4|$ with a p-value $p < 0.05$, and a Kendall strength $|\tau > 0.25|$ with a p-value $p < 0.05$. The correlations that satisfy both the strength and the statistical significance criteria defined for the Spearman and Kendall tests are reported on a heat map in Fig. \ref{fig:corr_heatmap} in terms of the average absolute Spearman-Kendall strength.

The QM metrics with the highest correlation with a human criteria (HM) are: Avg-min-distance-person, Proxemics-Intimate-Space, Proxemics-Social-Space, Time-to-goal, and Avg-robot-linear-speed. These results confirm the clustering analysis based on the cumulative ARI score (Fig. \ref{fig:histogram_cumulative_ARI}).

Social Work does not appear among the strongly correlating metrics, as we detected in Fig. \ref{fig:histogram_cumulative_ARI}. Although it is a complex social metric that combines distance and velocity via Social Forces between agents, it does not align with individual HM aspects. Proxemic scores and minimum distance to people correlate with unobtrusiveness, friendliness, and avoidance foresight. Average robot linear velocity relates only to avoidance capability, not to friendliness. Smoothness appears weakly captured by current metrics. While cumulative heading changes were expected to reflect this aspect, only time-to-goal shows sufficient correlation, suggesting it relates to multiple navigation aspects.

%An unexpected result is the absence of the Social Work in the table of strongly correlating metrics, which, among others, can be considered the most complete and advanced social metric, mixing distance and velocity effect according to the Social Forces generated by the interaction among agents. However, it does not match single HM aspects.
%The proxemic score and the minimum distance to people clearly correlate with unobtrusiveness, friendliness, and avoidance foresight. The average robot linear velocity is only tied to avoidance capability, but does not reflect the level of friendliness in the behavior. 
%Smoothness is a subtle aspect to evaluate, and it seems to be less represented by current metrics. The cumulative heading changes metric is expected to correlate more with that aspect. Instead, only time-to-goal shows a sufficient level of matching, showing that can be related to multiple aspects of navigation.

We supplemented the correlation study with a group-based statistical analysis using the Kruskal-Wallis H-test. % and Dunn’s post-hoc test. 
First, while Spearman and Kendall evaluate the monotonic relationship across all 24 experiments, the Kruskal-Wallis test assesses the discriminatory power of the metrics between the three different behavior levels: (Run 1, 2 and 3). %Second, this approach provides the statistical significance required to prove that the score gradients observed in the human data are mathematically meaningful, directly addressing the concern that apparent differences might fall within the margin of error.}

\begin{figure}[t]
    \centering
    \includegraphics[width=\linewidth]{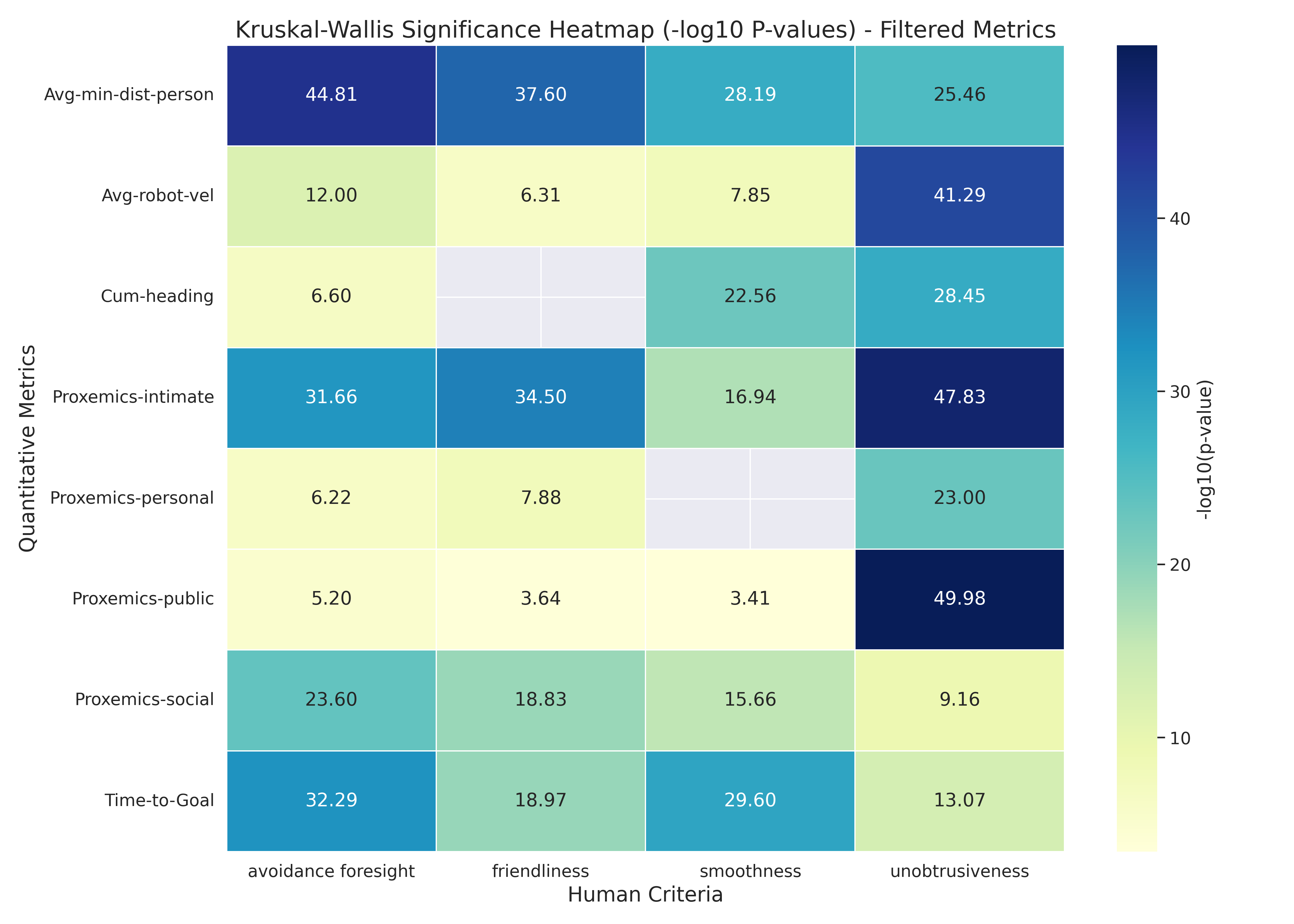}
    \caption{Kruskal Wallis significance heatmap (-$log_{10}$ $p-values$) illustrating the discriminatory power of quantitative metrics between the 3 experimental runs. Darker cells denote higher statistical confidence in a metric's ability to distinguish between social behavior level. The results confirm the clustering analysis.}
    \label{fig:kruskal_heatmap}
\end{figure}

The Kruskal-Wallis Significance Heatmap (Fig. \ref{fig:kruskal_heatmap}) illustrates the -$ log_{10}$ p-values, where higher values indicate greater statistical confidence in a metric's ability to distinguish between behavior runs. The results show that the metrics in our optimal QM$^*$ set, specifically Proxemics-intimate, Avg-min-dist-person, and Time-to-goal, demonstrate the highest significance across all human criteria. While Smoothness seems to be a subtle aspect poorly captured by metrics, according to both heatmaps, Unobtrusiveness reveal a shift from identifying simple monotonic relationships to establishing high-power statistical discriminators for robot behavior. Notably, the Social Work metric is too noisy what makes it a less reliable metric of social quality. Similarly, the Path length metric is also depreciable.
By triangulating these results with our clustering (ARI) and correlation (Spearman/Kendall) data, we provide a robust, human-aligned foundation for the QM$^*$ benchmarking protocol.

\subsection{Discussion and limitations}
The use of 2D top-down animations representing the environment allowed the survey participants to have a clear understanding of the velocity profiles, relative distances and path smoothness for the displayed scenarios. It should be noted that this methodology lacks the psychological factor of sharing a physical space with the robot.
%%%%%%%%%%%%%%%%%%%%%%%%%%%%%%%%%%%%%%%%%%%%%%%%%%%%%%%%%%%%
% CONCLUSIONS
%%%%%%%%%%%%%%%%%%%%%%%%%%%%%%%%%%%%%%%%%%%%%%%%%%%%%%%%%%%%
\section{Conclusions}
\label{sec:conclusions}
This work presents an in-depth analysis of social navigation metrics, investigating existing correlations between current quantitative metrics and qualitative human assessment, with the aim of exploiting hidden data patterns to provide automated benchmarks that align with human judgment. The study proposed a clustering method to identify a QM feature space where experiments are grouped. Then, a statistical analysis of correlation have been performed to inspect individual links between metrics. Our findings show that by selecting relevant QM sets from the correlation results, it is possible to obtain an overall experimental benchmarking that correlates with human-like assessment, aiding in the preliminary screening of social navigation algorithms. Indeed, optimal QM* sets demonstrate clearer scoring trends and behavior identification compared to the complete ensemble of QM. Nonetheless, advanced and mixed social scenarios still represent a challenge due to the complexity of correctly evaluating a navigation experiment that is only partially acceptable, leaving room for future improvement.
Although the study identifies strong correlations with human perception, we acknowledge that human surveys remain the gold standard for final validation. The current study is limited by a sample size of 24 real experimental trajectories; therefore, the study proposes a complementary benchmarking tool to human-centered evaluation.
% \textcolor{blue}{Another aspect to consider is that providing metric intuition may have introduced framing bias. Even though it is done to ensure consistent interpretation across participants, future work will investigate if unaided human evaluations yield similar results.}
Future works will build on these findings to further highlight hidden relationships between QM and HM in larger and mixed datasets. Learning and regression methods will be investigated with the aim of inferring human assessment from quantitative metrics, particularly focusing on the importance and explainability of characteristics. 

\addtolength{\textheight}{-12cm}   % This command serves to balance the column lengths

%%%%%%%%%%%%%%%%%%%%%%%%%%%%%%%%%%%%%%%%%%%%%%%%%%%%%%%%%%%%%%%%%%%%%%%%%%%%%%%%

%%%%%%%%%%%%%%%%%%%%%%%%%%%%%%%%%%%%%%%%%%%%%%%%%%%%%%%%%%%%%%%%%%%%%%%%%%%%%%%%

%%%%%%%%%%%%%%%%%%%%%%%%%%%%%%%%%%%%%%%%%%%%%%%%%%%%%%%%%%%%%%%%%%%%%%%%%%%%%%%%

\bibliographystyle{IEEEtran} %alpha
\bibliography{biblio}

@article{moller2021survey,
  title={A survey on human-aware robot navigation},
  author={M{\"o}ller, Ronja and Furnari, Antonino and Battiato, Sebastiano and H{\"a}rm{\"a}, Aki and Farinella, Giovanni Maria},
  journal={Robotics and Autonomous Systems},
  volume={145},
  pages={103837},
  year={2021},
  publisher={Elsevier}
}

@article{gao2022evaluation,
  title={Evaluation of socially-aware robot navigation},
  author={Gao, Yuxiang and Huang, Chien-Ming},
  journal={Frontiers in Robotics and AI},
  volume={8},
  pages={721317},
  year={2022},
  publisher={Frontiers}
}

@inproceedings{singamaneni2023towards,
  title={Towards benchmarking human-aware social robot navigation: A new perspective and metrics},
  author={Singamaneni, Phani Teja and Favier, Anthony and Alami, Rachid},
  booktitle={2023 32nd IEEE International Conference on Robot and Human Interactive Communication (RO-MAN)},
  pages={914--921},
  year={2023},
  organization={IEEE}
}

@article{samarakoon2022review,
  title={A review on human--robot proxemics},
  author={Samarakoon, SM Bhagya P and Muthugala, MA Viraj J and Jayasekara, AG Buddhika P},
  journal={Electronics},
  volume={11},
  number={16},
  pages={2490},
  year={2022},
  publisher={MDPI}
}

@inproceedings{wang2022metrics,
  title={Metrics for evaluating social conformity of crowd navigation algorithms},
  author={Wang, Junxian and Chan, Wesley P and Carreno-Medrano, Pamela and Cosgun, Akansel and Croft, Elizabeth},
  booktitle={2022 IEEE International Conference on Advanced Robotics and Its Social Impacts (ARSO)},
  pages={1--6},
  year={2022},
  organization={IEEE}
}

@article{stratton2024characterizing,
  title={Characterizing the Complexity of Social Robot Navigation Scenarios},
  author={Stratton, Andrew and Hauser, Kris and Mavrogiannis, Christoforos},
  journal={IEEE Robotics and Automation Letters},
  year={2024},
  publisher={IEEE}
}

@article{mavrogiannis2023core,
  title={Core challenges of social robot navigation: A survey},
  author={Mavrogiannis, Christoforos and Baldini, Francesca and Wang, Allan and Zhao, Dapeng and Trautman, Pete and Steinfeld, Aaron and Oh, Jean},
  journal={ACM Transactions on Human-Robot Interaction},
  volume={12},
  number={3},
  pages={1--39},
  year={2023},
  publisher={ACM New York, NY}
}

@ARTICLE{karwowski23metrics,
  author={Karwowski, Jarosław and Szynkiewicz, Wojciech},
  journal={IEEE Access}, 
  title={Quantitative Metrics for Benchmarking Human-Aware Robot Navigation}, 
  year={2023},
  volume={11},
  number={},
  pages={79941-79953},
  doi={10.1109/ACCESS.2023.3299178}}

@ARTICLE{PerezRal2023,
  author={Pérez-Higueras, Noé and Otero, Roberto and Caballero, Fernando and Merino, Luis},
  journal={IEEE Robotics and Automation Letters}, 
  title={HuNavSim: A ROS 2 Human Navigation Simulator for Benchmarking Human-Aware Robot Navigation}, 
  year={2023},
  month={September},
  volume={8},
  number={11},
  pages={7130-7137},
  issn={2377-3766},
  doi={10.1109/LRA.2023.3316072}}

@article{biswas2022socnavbench,
  title={Socnavbench: A grounded simulation testing framework for evaluating social navigation},
  author={Biswas, Abhijat and Wang, Allan and Silvera, Gustavo and Steinfeld, Aaron and Admoni, Henny},
  journal={ACM Transactions on Human-Robot Interaction (THRI)},
  volume={11},
  number={3},
  pages={1--24},
  year={2022},
  publisher={ACM New York, NY}
}

@article{helbing1995social,
  title={Social force model for pedestrian dynamics},
  author={Helbing, Dirk and Molnar, Peter},
  journal={Physical review E},
  volume={51},
  number={5},
  pages={4282},
  year={1995},
  publisher={APS}
}

@article{hirose2024selfi,
  title={Selfi: Autonomous self-improvement with reinforcement learning for social navigation},
  author={Hirose, Noriaki and Shah, Dhruv and Stachowicz, Kyle and Sridhar, Ajay and Levine, Sergey},
  journal={arXiv preprint arXiv:2403.00991},
  year={2024}
}

@article{likas2003global,
  title={The global k-means clustering algorithm},
  author={Likas, Aristidis and Vlassis, Nikos and Verbeek, Jakob J},
  journal={Pattern recognition},
  volume={36},
  number={2},
  pages={451--461},
  year={2003},
  publisher={Elsevier}
}

@article{steinley2004properties,
  title={Properties of the hubert-arable adjusted rand index.},
  author={Steinley, Douglas},
  journal={Psychological methods},
  volume={9},
  number={3},
  pages={386},
  year={2004},
  publisher={American Psychological Association}
}

@article{Francis2025,
author = {Francis, Anthony and P\'{e}rez-D’Arpino, Claudia and Li, Chengshu and Xia, Fei and Alahi, Alexandre and Alami, Rachid and Bera, Aniket and Biswas, Abhijat and Biswas, Joydeep and Chandra, Rohan and Chiang, Hao-Tien Lewis and Everett, Michael and Ha, Sehoon and Hart, Justin and How, Jonathan P. and Karnan, Haresh and Lee, Tsang-Wei Edward and Manso, Luis J. and Mirsky, Reuth and Pirk, S\"{o}ren and Singamaneni, Phani Teja and Stone, Peter and Taylor, Ada V. and Trautman, Peter and Tsoi, Nathan and V\'{a}zquez, Marynel and Xiao, Xuesu and Xu, Peng and Yokoyama, Naoki and Toshev, Alexander and Mart\'{\i}n-Mart\'{\i}n, Roberto},
title = {Principles and Guidelines for Evaluating Social Robot Navigation Algorithms},
year = {2025},
issue_date = {June 2025},
publisher = {Association for Computing Machinery},
address = {New York, NY, USA},
volume = {14},
number = {2},
url = {https://doi.org/10.1145/3700599},
doi = {10.1145/3700599},
journal = {J. Hum.-Robot Interact.},
month = feb,
articleno = {34},
numpages = {65}
}

@book{proxemics_Hall,
    author = {Hall, Edward T.},
    howpublished = {Paperback},
    isbn = {0385084765},
    month = oct,
    publisher = {Anchor},
    title = {{The Hidden Dimension}},
    year = {1990}
}

@ARTICLE{sean2_Tsoi_RAL22,
  author={Tsoi, Nathan and Xiang, Alec and Yu, Peter and Sohn, Samuel S. and Schwartz, Greg and Ramesh, Subashri and Hussein, Mohamed and Gupta, Anjali W. and Kapadia, Mubbasir and Vázquez, Marynel},
  journal={IEEE Robotics and Automation Letters}, 
  title={SEAN 2.0: Formalizing and Generating Social Situations for Robot Navigation}, 
  year={2022},
  volume={7},
  number={4},
  pages={11047-11054},
  doi={10.1109/LRA.2022.3196783}
}

@article{Arena4_2024,
    author = {Volodymyr and Kästner, Linh and Diaz, Diego and Nguyen, Huu and Schreff, Maximilian and Lenz, Tim and Kreutz, Jonas and Martban, Ahmed and Zeng, Huajian and Soh, Harold},
    year = {2024},
    month = {09},
    pages = {},
    journal = {ArXiv pre-print},
    title = {Arena 4.0: A Comprehensive ROS2 Development and Benchmarking Platform for Human-centric Navigation Using Generative-Model-based Environment Generation},
    doi = {10.48550/arXiv.2409.12471}
}

@article{eirale2025human,
  title={Human following and guidance by autonomous mobile robots: A comprehensive review},
  author={Eirale, Andrea and Martini, Mauro and Chiaberge, Marcello},
  journal={IEEE Access},
  year={2025},
  publisher={IEEE}
}

@article{Friedman1937,
 ISSN = {01621459, 1537274X},
 URL = {http://www.jstor.org/stable/2279372},
 author = {Milton Friedman},
 journal = {Journal of the American Statistical Association},
 number = {200},
 pages = {675--701},
 publisher = {[American Statistical Association, Taylor & Francis, Ltd.]},
 title = {The Use of Ranks to Avoid the Assumption of Normality Implicit in the Analysis of Variance},
 urldate = {2026-03-27},
 volume = {32},
 year = {1937}
}

@article{wilcoxon1945,
 ISSN = {00994987},
 URL = {http://www.jstor.org/stable/3001968},
 author = {Frank Wilcoxon},
 journal = {Biometrics Bulletin},
 number = {6},
 pages = {80--83},
 publisher = {[International Biometric Society, Wiley]},
 title = {Individual Comparisons by Ranking Methods},
 urldate = {2026-03-27},
 volume = {1},
 year = {1945}
}

@INPROCEEDINGS{Martini2024-SFWP,
  author={Martini, Mauro and Pérez-Higueras, Noé and Ostuni, Andrea and Chiaberge, Marcello and Caballero, Fernando and Merino, Luis},
  booktitle={2024 IEEE/RSJ International Conference on Intelligent Robots and Systems (IROS)}, 
  title={Adaptive Social Force Window Planner with Reinforcement Learning}, 
  year={2024},
  volume={},
  number={},
  pages={4816-4822},
  doi={10.1109/IROS58592.2024.10802383}}

\end{document}